\documentclass[conference]{IEEEtran}
\IEEEoverridecommandlockouts

\usepackage{cite}
\usepackage{amsmath,amssymb,amsfonts}
\usepackage{algorithm}
\usepackage{algpseudocode}

\usepackage{booktabs}
\usepackage{multirow}
\usepackage{graphicx}
\usepackage{textcomp}
\usepackage{xcolor}
\usepackage{url} 

\def\BibTeX{{\rm B\kern-.05em{\sc i\kern-.025em b}\kern-.08em
    T\kern-.1667em\lower.7ex\hbox{E}\kern-.125emX}}
    \usepackage{tikz}
\usetikzlibrary{shapes.geometric, arrows}
\begin{document}

\title{\textsc{RoSum-Mcts}: Monte Carlo Tree Search-Inspired HDL Code Summarization with Structural Rewards }


\author{\IEEEauthorblockN{Prashanth Vijayaraghavan}
\IEEEauthorblockA{
\textit{IBM Research}\\
San Jose, CA, USA \\
prashanthv@ibm.com}
\and
\IEEEauthorblockN{Charles Mackin}
\IEEEauthorblockA{
\textit{IBM Research}\\
San Jose, CA, USA \\
charles.mackin@ibm.com}
\and
\IEEEauthorblockN{Luyao Shi}
\IEEEauthorblockA{
\textit{IBM Research}\\
San Jose, CA, USA \\
luyao.shi@ibm.com}
\and
\IEEEauthorblockN{Apoorva Nitsure}
\IEEEauthorblockA{
\textit{IBM Research}\\
San Jose, CA, USA \\
Apoorva.Nitsure@ibm.com}
\and
\IEEEauthorblockN{Ashutosh Jadhav}
\IEEEauthorblockA{
\textit{IBM Research}\\
San Jose, CA, USA \\
ashutosh@us.ibm.com}

\and
\IEEEauthorblockN{David Beymer}
\IEEEauthorblockA{
\textit{IBM Research}\\
San Jose, CA, USA \\
beymer@us.ibm.com}
\and
\IEEEauthorblockN{Tyler Baldwin}
\IEEEauthorblockA{
\textit{IBM Research}\\
San Jose, CA, USA \\
tbaldwin@us.ibm.com
}
\and
\IEEEauthorblockN{Ehsan Degan}
\IEEEauthorblockA{
\textit{IBM Research}\\
San Jose, CA, USA \\
edehgha@us.ibm.com
}
\and
\IEEEauthorblockN{Vandana Mukherjee}
\IEEEauthorblockA{
\textit{IBM Research}\\
San Jose, CA, USA \\
vandana@us.ibm.com
}
}

\maketitle


\begin{abstract}  
Large language models (LLMs) have shown promise in code summarization, yet their effectiveness for Hardware Description Languages (HDLs) like VHDL and Verilog remains underexplored. We propose \textsc{RoSum-Mcts}, an LLM-guided approach inspired by Monte Carlo Tree Search (MCTS) that refines summaries through structured exploration and reinforcement-driven optimization. Our method integrates both local and global context via a hierarchical candidate expansion mechanism and optimizes summaries using a composite reward function balancing functional correctness (FC), local content adequacy (LCA), and fluency. We evaluate \textsc{RoSum-Mcts} on the VHDL-eval and Verilog-eval datasets, demonstrating its consistent outperformance over baseline methods by leveraging structured bottom-up refinement and reinforcement-based optimization. Ablation studies confirm the necessity of both local and global expansion strategies, as well as the importance of balancing FC and LCA for optimal performance. Furthermore, \textsc{RoSum-Mcts} proves robust against superficial modifications, such as variable renaming, maintaining summary quality where baselines degrade. These results establish \textsc{RoSum-Mcts} as an effective and robust HDL summarization framework, paving the way for further research into reinforcement-enhanced code summarization.  
\end{abstract}


\section{Introduction}

Large Language Models (LLMs) have significantly advanced code-related tasks, including generation, translation, and summarization, thereby enhancing software development and maintenance processes \cite{fakih2024llm4plc,paul2024ircoder,wei2023magicoder,yang2023harnessing,yang2023harnessing,cassano2024knowledge,li2023starcoder, roziere2023code}. However, their application to Hardware Description Languages (HDLs) such as VHDL and Verilog has been relatively limited and has specifically been focused more on Verilog \cite{zhao2024codev,cui2024origen,gao2024autovcoder,pei2024betterv}. HDLs present unique challenges due to their inherent concurrency, hardware-specific semantics, and structural complexity, which are not typically encountered in general-purpose programming languages. Recent studies have begun to explore the application of LLMs in HDL code generation and summarization. For instance, recent works, such as AutoChip, have investigated automating HDL generation using LLM feedback, focusing on improving functional correctness but not specifically targeting summarization tasks \cite{thakur2023autochip}. Similarly, studies on classification-based and paradigm-based automatic HDL code generation using LLMs have aimed to mitigate issues like hallucination in code generation but have not directly addressed the robustness of code summarization \cite{sun2024classification}. 

In the area focused on HDL code summarization, the \textsc{CoDes} approach leverages Abstract Syntax Trees (ASTs) to generate intermediate descriptive information and use that to generate the overall summary of the VHDL code \cite{vijayaraghavan2024chain}. While \textsc{CoDes} utilizes structural insights from ASTs to produce summaries, it does not explicitly address robustness against syntactic variations like identifier renaming.  A notable challenge in HDL code summarization is the sensitivity of LLMs to superficial syntactic variations, such as changes in variable names, module identifiers, or signal names. These variations can lead to inconsistencies in generated summaries, as LLMs may misinterpret or overlook the core functionality of the code when faced with different naming conventions. This issue highlights the necessity for summarization approaches that are robust to such variations, ensuring accurate and consistent representation of the code's intent.

To address these challenges, we introduce \textsc{RoSum-Mcts}, a novel approach inspired by Monte Carlo Tree Search (MCTS) \cite{li2024rethinkmcts,dainese2024generating,browne2012survey}, traditionally used in sequential decision-making tasks like game playing. In the context of HDL code summarization, MCTS principles can be adapted for structured exploration and optimization. Our method integrates both local and global contexts through a hierarchical candidate expansion mechanism, systematically exploring and evaluating potential summary refinements. Unlike conventional summarization methods that rely solely on sequence-to-sequence modeling, \textsc{RoSum-Mcts} incorporates reinforcement-driven optimization to iteratively refine initial LLM-generated summaries. The refinement process is guided by a composite reward function that balances multiple objectives: functional correctness (FC), ensuring the generated summary accurately reflects the core functionality of the HDL design; local content adequacy (LCA), measuring whether critical design aspects are preserved in the summary; and fluency, ensuring the generated summary remains readable and grammatically coherent. By optimizing these objectives through reinforcement learning, \textsc{RoSum-Mcts} produces more robust and informative summaries compared to existing approaches.  We conduct extensive experiments on both VHDL and Verilog datasets \cite{liu2023verilogeval,vijayaraghavan2024vhdl}, evaluating the effectiveness of \textsc{RoSum-Mcts} across multiple LLMs and comparing it with state-of-the-art baseline methods. Our results demonstrate that \textsc{RoSum-Mcts} consistently outperforms existing summarization techniques, achieving superior results across various automatic evaluation metrics. Our key contributions in this work are as follows: 

\noindent \textbf{MCTS-inspired Summarization:} 
We introduce \textsc{RoSum-Mcts}, an MCTS-inspired approach for HDL code summarization that refines LLM-generated summaries through structured exploration and reinforcement-driven optimization, integrating both local and global contexts via a hierarchical candidate expansion mechanism.

\noindent \textbf{Comprehensive Evaluation:} 
Extensive experiments on VHDL and Verilog datasets show that \textsc{RoSum-Mcts} remains effective under perturbations like variable/module renaming, outperforming baselines across LLMs and metrics.

\section{Related Work}
Large Language Models (LLMs) have significantly improved code summarization by generating concise and informative summaries across multiple languages. Sun et al. \cite{sun2024source} provide a detailed analysis of LLMs' summarization capabilities, demonstrating their effectiveness. However, LLMs remain sensitive to variations in variable names and module identifiers, leading to inconsistencies in generated summaries. HDLs such as VHDL and Verilog pose challenges for summarization due to their concurrency and hardware-specific semantics. Traditional methods often struggle with these complexities, necessitating specialized approaches. The \textsc{CoDes} framework \cite{vijayaraghavan2024chain} enhances VHDL summarization using intermediate descriptions, while \textsc{CodeV} \cite{zhao2024codev} improves Verilog summarization through multi-level summarization. Despite these advances, existing methods remain sensitive to superficial changes in variable and module names, impacting summary quality. To address these limitations, we propose \textsc{RoSum-Mcts}, an MCTS-inspired approach that refines LLM-generated summaries through structured exploration. By integrating local and global contexts via hierarchical candidate expansion and optimizing functional correctness, local content adequacy, and fluency, \textsc{RoSum-Mcts} enhances robustness against syntactic variations. Our experiments across multiple datasets and LLMs show that \textsc{RoSum-Mcts} consistently outperforms existing methods, setting a new standard for HDL code summarization.

\section{Methodology}

\begin{figure*}[h]
    \centering
    \includegraphics[width=0.6\textwidth]{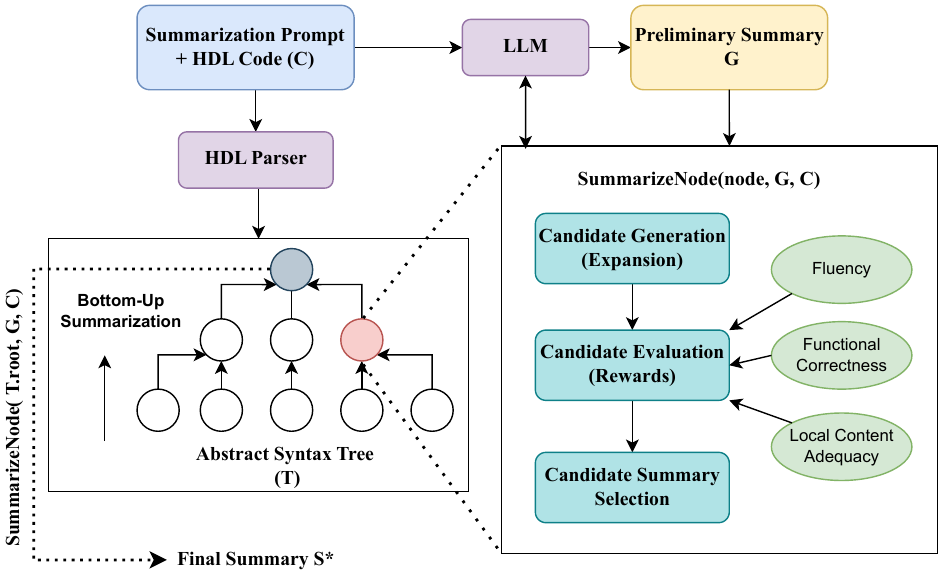}
    \caption{Illustration of our \textsc{RoSum-Mcts} method for HDL Code Summarization. }
    \label{fig:architecture}
\end{figure*}

In this section, we describe the proposed \textsc{RoSum-Mcts} framework for HDL code summarization. \textsc{RoSum-Mcts} is a hierarchical method that integrates Abstract Syntax Trees (ASTs) with Large Language Models (LLMs) and is inspired by Monte Carlo Tree Search (MCTS). Rather than applying MCTS directly to sequential decision-making tasks, our approach adapts the algorithm for HDL code summarization through a bottom-up strategy. In this framework, summaries are incrementally refined from the leaf nodes upward by leveraging prompt-based candidate generation and a composite reward evaluation. This design preserves the structural and semantic integrity of the HDL design, producing final summaries that are robust to superficial variations such as changes in variable or module names. Figure \ref{fig:architecture} depicts the \textsc{RoSum-Mcts} framework for HDL code summarization.

\subsection{HDL Parsing and AST Construction}
To capture both the structural and behavioral aspects of HDL designs, we first parse the HDL file \(F\) to construct its AST. For Verilog HDL, we utilize \texttt{pyverilog}---a Python toolkit designed for RTL design analysis and code generation \cite{Takamaeda:2015:ARC:Pyverilog}. This toolkit efficiently extracts ASTs that reflect the design’s structure and functionality. In the case of VHDL, \texttt{pyVHDLParser} is employed---a streaming-based parser that tokenizes VHDL-2008 source files and constructs detailed ASTs \cite{pyVHDLParser}. Although our initial focus is on single-file parsing, both tools support cross-file dependency resolution, ensuring that inter-file references are accurately maintained in the AST.

\subsection{Generation of Preliminary Global Summary}
Before initiating the hierarchical summarization process, a preliminary summary \(G\) of the entire HDL code is generated using an LLM-based summarization approach. This global summary provides essential context and guides the candidate generation process at each node, thereby enhancing the quality and variability of the generated node-level summaries.

\subsection{MCTS-Inspired Hierarchical HDL Code Summarization Approach}
\subsubsection{Overview and Motivation}
Traditional MCTS is widely used in sequential decision-making tasks, such as game playing, by exploring multiple candidate paths via simulation rollouts. In \textsc{RoSum-Mcts}, we adapt this idea to HDL code summarization through a bottom-up approach. At each non-leaf node in the AST, multiple candidate summaries are generated using various prompt templates. These candidates are evaluated using a composite reward function that integrates a \emph{fluency reward}, a \emph{local content adequacy reward}, and a \emph{functional correctness reward} (the detailed reward design is provided in Section~\ref{sec:reward_function_design}). The candidate with the highest composite score is then selected as the refined summary for that node. This one-step candidate expansion and direct evaluation (rather than full rollouts) improves computational efficiency while preserving the structural and semantic integrity of the HDL design, ultimately producing a final summary that is robust to naming perturbations.

\subsubsection{Procedure}
The summarization process in \textsc{RoSum-Mcts} proceeds through the following steps:

\begin{itemize}
    \item \textbf{Parsing and Global Context:} The HDL file is parsed using \texttt{pyverilog} (for Verilog) or \texttt{pyVHDLParser} (for VHDL) to construct the AST. Simultaneously, a preliminary global summary of the entire code is generated using an LLM-based summarization approach, which provides essential context for guiding subsequent summarization tasks.
    
    \item \textbf{Recursive Summarization:} The algorithm traverses the AST in a bottom-up manner. For each node, summaries of its immediate children are first generated. If a node is a leaf, its summary is produced directly; otherwise, the child summaries are aggregated to form an ordered list.
    
    \item \textbf{Candidate Generation:} At every non-leaf node, multiple candidate summaries are generated using various prompt templates (see Table~\ref{tab:prompt_variants} in Appendix). These candidates leverage both the local child summaries and the global context to capture different facets of the node’s functionality.
    
    \item \textbf{Candidate Evaluation and Selection:} Each candidate summary is evaluated using a composite reward function that integrates the \emph{fluency reward}, the \emph{local content adequacy reward}, and the \emph{functional correctness reward} (the detailed formulation is provided in Section~\ref{sec:reward_function_design}). The candidate with the highest composite reward is selected as the refined summary for the node.
    
    \item \textbf{Propagation:} The refined summary is propagated upward in the AST, ensuring that each higher-level node benefits from the best local information, ultimately leading to a robust final summary.
\end{itemize}

\subsubsection{Candidate Expansion Mechanism}
At every non-leaf AST node, candidate expansion is achieved using four distinct prompt templates (see Table~\ref{tab:prompt_variants}). These templates are designed to capture various dimensions of the code's functionality:
\begin{itemize}
    \item \textbf{Type 1 (Local Summary Context):} This prompt combines the child summaries to generate a concise, localized summary that reflects the immediate details of the node.
    \item \textbf{Type 2 (Local Inference):} Here, the focus is solely on the child summaries to infer the higher-level functionality, abstracting the underlying behavior of the node.
    \item \textbf{Type 3 (Global Code Context):} This variant integrates the full code context along with the child summaries, ensuring that the generated summary is informed by the complete source code.
    \item \textbf{Type 4 (Global Summary Context):} This template leverages the preliminary global summary together with the child summaries to refine the description of the current node’s functionality.
\end{itemize}

These candidate expansion techniques are crucial as they ensure that both the local details and the overall global context of the HDL code are captured, ultimately enhancing the coherence and completeness of the final summary. Table~\ref{tab:prompt_variants} presents the prompt templates used to generate candidate summaries.

\subsubsection{Reward Function Design}
\label{sec:reward_function_design}

In our hierarchical HDL code summarization framework, the quality of a candidate summary \(S_i\) is evaluated by integrating three distinct reward components. Each component is designed to assess a key aspect of summary quality—ensuring fluency, content adequacy, and functional correctness.

\paragraph{Fluency Reward \(R_F(S_i)\)}
Fluency is crucial for producing summaries that are readable and easily understood by developers. We quantify fluency by measuring the perplexity of the candidate summary, \(\text{PPL}(S_i)\). The fluency reward is then defined as:
\[
R_F(S_i) = \frac{1}{1 + \text{PPL}(S_i)},
\]
which maps lower perplexity values (indicating higher fluency) to scores closer to 1.

\paragraph{Local Content Adequacy Reward \(R_{LCA}(S_i, S_{\text{child\_list}})\)}
This reward evaluates whether the candidate summary accurately and comprehensively encapsulates the ordered sequence of its child summaries. Let
\[
S_{\text{child\_list}} = \bigl[ S_{\text{child}_1}, S_{\text{child}_2}, \dots, S_{\text{child}_k} \bigr]
\]
represent the ordered list of children summaries.

\begin{algorithm}[H]
\caption{\textsc{RoSum-MCTS}: Hierarchical HDL Code Summarization}
\label{alg:RoSum-MCTS-Sequential}
\begin{algorithmic}

\State \textbf{Input:} HDL file $F$
\State \textbf{Output:} Final summary $S^\ast$

\State \textit{// Parse HDL file into AST}
\State $T \gets \textproc{ParseHDLFile}(F)$

\State \textit{// Load HDL source code}
\State $C \gets \textproc{LoadHDLFile}(F)$

\State \textit{// Generate global context summary}
\State $G \gets \textproc{PreliminarySummary}(C)$

\Function{SummarizeNode}{node, G, C}

    \If{\Call{IsLeaf}{node}}

        \State \textit{// Generate summary for leaf node}
        \State $S_{node} \gets \textproc{GenerateSummary}(node)$
        \State \Return $S_{node}$

    \Else

        \State \textit{// Initialize child summary list}
        \State $S_{\text{child\_list}} \gets [\,]$

        \For{$j = 1$ to $|node.\text{children}|$}

            \State $child_j \gets node.\text{children}[j]$
            \State $S_{\text{child}_j}
                   \gets \Call{SummarizeNode}{child_j, G, C}$

            \State Append $S_{\text{child}_j}$ to $S_{\text{child\_list}}$

        \EndFor

        \State \textit{// Candidate expansion}
        \State $\mathcal{S} \gets \textproc{GenerateCandidates}(S_{\text{child\_list}}, G, C)$

        \For{each candidate $S_i \in \mathcal{S}$}

            \State $R(S_i) \gets
                   \alpha R_F(S_i)
                   + \beta R_{LCA}(S_i,S_{\text{child\_list}})
                   + \gamma R_{FC}(S_i,C)$

        \EndFor

        \State \textit{// Select best candidate}
        \State $S_{node} \gets \arg\max_{S_i \in \mathcal{S}} R(S_i)$

        \State \Return $S_{node}$

    \EndIf

\EndFunction

\State \textit{// Main execution}
\State $S^\ast \gets \Call{SummarizeNode}{T.\text{root}, G, C}$

\State \Return $S^\ast$

\end{algorithmic}
\end{algorithm}

To assess content adequacy, we compute semantic embeddings of each child summary using SimCSE~\cite{gao2021simcse} and average them as:

\[
\bar{E}(S_{\text{child\_list}}) = \frac{1}{k} \sum_{j=1}^{k} E_{\text{SimCSE}}\bigl(S_{\text{child}_j}\bigr).
\]
We next determine the cosine similarity between the embedding of the candidate summary, \(E_{\text{SimCSE}}(S_i)\), and the aggregated embedding \(\bar{E}(S_{\text{child\_list}})\). Scaling this similarity to the range \([0,1]\) yields:
\[
R_{LCA}(S_i, S_{\text{child\_list}}) = \frac{\cos\Bigl(E_{\text{SimCSE}}(S_i),\, \bar{E}(S_{\text{child\_list}})\Bigr) + 1}{2}.
\]

\paragraph{Functional Correctness Reward \(R_{FC}(S_i, C)\)}
This reward verifies whether the candidate summary \(S_i\) accurately describes a submodule that is functionally present within the complete HDL code \(C\). Instead of generating an intermediate global summary, we directly employ a prompt-based evaluation using an LLM. The LLM is provided with the full HDL code \(C\) along with \(S_i\) and is tasked with assessing the consistency of \(S_i\) with the functionality of \(C\). The LLM returns a normalized score (between 0 and 1) as follows:
\[
R_{FC}(S_i, C) = \texttt{LLM\_Score}\Bigl(\texttt{Prompt}(C, S_i)\Bigr).
\]

\paragraph{Composite Reward}
The overall composite reward for a candidate summary is computed as a weighted sum of the three components:
\[
R(S_i) = \alpha \cdot R_F(S_i) + \beta \cdot R_{LCA}(S_i, S_{\text{child\_list}}) + \gamma \cdot R_{FC}(S_i, C),
\]
where \(\alpha\), \(\beta\), and \(\gamma\) are hyperparameters that balance the contributions of fluency, local content adequacy, and functional correctness, respectively. This composite reward function ensures that the selected summary is fluent, maintains submodule structure, and aligns functionally with the HDL code. The complete process is detailed in Algorithm \ref{alg:RoSum-MCTS-Sequential}.

\section{Experiments}

\subsection{Overview}
In this section, we outline our experimental setup to assess the effectiveness of our \textsc{RoSum-Mcts} method for HDL code summarization. Our experiments are designed to answer the following research questions:

\noindent \textbf{RQ1:} How effective is our \textsc{RoSum-Mcts} method compared to existing baseline approaches?

\noindent  \textbf{RQ2:} How do different aspects of \textsc{RoSum-Mcts} (e.g., candidate expansion mechanism, composite rewards, and hyperparameters) contribute to the overall performance?

\noindent  \textbf{RQ3:} How robust is the approach to superficial modifications, such as changes in variable or module names?

\subsection{Datasets}
We evaluate our approach on two widely used HDL datasets:
(a) \textbf{Verilog-Eval:} A benchmark suite for Verilog HDL code summarization consisting of 156 problems sourced from the HDLBits website \cite{liu2023verilogeval}; and (b) \textbf{VHDL-Eval:} A comprehensive dataset for VHDL code summarization comprising 202 code problems tailored for VHDL \cite{vijayaraghavan2024vhdl}.

\begin{table*}[h]
    \centering
    \caption{Comparison of summarization approaches across different models and datasets. Higher values indicate better performance.}
    \label{tab:results}
    \begin{tabular}{lccccccccc}
        \toprule
        \multirow{2}{*}{Dataset} & \multirow{2}{*}{Method} & \multicolumn{2}{c}{GPT-4o} & \multicolumn{2}{c}{Llama-3} & \multicolumn{2}{c}{Granite Code-34b} & \multicolumn{2}{c}{DeepSeek-Coder-33b} \\
        \cmidrule(lr){3-4} \cmidrule(lr){5-6} \cmidrule(lr){7-8} \cmidrule(lr){9-10}
        & & LLM PR & Rouge-L & LLM PR & Rouge-L & LLM PR & Rouge-L & LLM PR & Rouge-L \\
        \midrule
        \multirow{3}{*}{VHDL}   & Vanilla        & 48.7 & 39.3 & 42.6 & 34.1 & 34.2 & 29.5 & 26.8 & 25.1 \\
                                & \textsc{CoDes}          & 52.3 & 43.2 & 46.1 & 39.5 & 39.1 & 33.4 & 29.4 & 28.6 \\
                                & \textsc{RoSum-Mcts} & \textbf{60.9} & \textbf{50.8} & 53.2 & 45.9 & 45.6 & 39.3 & 35.7 & 33.1 \\
        \midrule
        \multirow{3}{*}{Verilog} & Vanilla        & 52.4 & 43.1 & 45.9 & 40.2 & 40.5 & 33.0 & 32.1 & 29.5 \\
                                & \textsc{CoDes}          & 55.0 & 46.9 & 48.7 & 43.5 & 43.8 & 37.8 & 35.5 & 31.2 \\
                                & \textsc{RoSum-Mcts} & \textbf{64.2} & \textbf{55.4} & 55.6 & 48.8 & 50.3 & 44.2 & 40.5 & 36.7 \\
        \bottomrule
    \end{tabular}
\end{table*}

\subsection{Metrics}
We employ the following automatic metrics to assess the quality of the generated summaries: (a)  \textbf{ROUGE-L:} Measures the longest common subsequence between the generated summary and the reference summary, emphasizing fluency and coherence \cite{Lin:2004:ROUGE}; and (b) \textbf{LLM Preferred Rate (PR):} Utilizes large language models (LLMs) to rate and compare the quality of summaries, serving as a proxy for human judgment \cite{yuan2023evaluating}.

\subsection{Baseline Models and Approaches}
For evaluation, we consider several LLMs which includes proprietary and open source models such as GPT-4o, Llama-3, DeepSeek-Coder-33b, Granite Code-34b, and CodeLlama 34b. To benchmark our method, we compare against two baseline approaches:
(a) \textbf{Vanilla Prompting:} A straightforward approach where a simple prompt is used to summarize the HDL code; and (b) \textbf{\textsc{CoDes}:} A multi-step prompting strategy, referred to as Chain-of-Description, that aggregates summaries from different components of the AST. We evaluate \textsc{RoSum-Mcts} alongside baselines to demonstrate its strength in capturing local details and global code context.


\subsection{Implementation Details and Experimental Setup}

Our experimental framework is implemented in PyTorch \cite{paszke2019pytorch}, using ChatGPT APIs \cite{openai2023chatgpt} for generating global and candidate summaries. The system runs on two NVIDIA V100 GPUs (32 GB each), supporting high-throughput inference and embedding computations via models like SimCSE \cite{gao2021simcse}. Candidate generation and reward evaluation are batched to reduce latency and improve GPU utilization. Parallel processing and asynchronous execution maximize hardware efficiency. Reward weights (\(\alpha, \beta, \gamma\)) are tuned on validation data. With an average of $\sim{62.7}$ AST nodes per module and $\sim{290}$ tokens per end-to-end prompt across evaluation datasets, our batching, pruning, and chunking strategies minimize both LLM calls and token cost. These optimizations ensure that \textsc{RoSum-Mcts} remains scalable and cost-efficient across both local and proprietary model deployments.


\section{Results}

\subsection{(RQ1) Effectiveness of our \textsc{RoSum-Mcts} Method}

Table~\ref{tab:results} presents the evaluation of our proposed \textsc{RoSum-Mcts} method compared to the baseline approaches—Vanilla and Codes—across two datasets (VHDL and Verilog) using various large language models. We report results using LLM Preferred Rate (LLM PR) and Rouge-L as key performance metrics. Table~\ref{tab:results} clearly demonstrates that \textsc{RoSum-Mcts} consistently outperforms the baseline methods across all datasets and models. GPT-4o achieves the highest LLM PR and Rouge-L scores, indicating superior summarization capability. Llama-3 follows closely, surpassing Granite Code-34b, which in turn performs better than DeepSeek-Coder-33b. Among code-specific LLMs, Granite Code-34b exhibits strong performance relative to DeepSeek-Coder-33b and CodeLlama-34b. Since DeepSeek-Coder-33b and CodeLlama-34b yield comparable results with only minor variations, we report only one of them in the table. Verilog consistently outperforms VHDL in summarization quality, likely due to the increased exposure of LLMs to Verilog data compared to VHDL, given Verilog's greater popularity and wider adoption. Although the \textsc{CoDes} method employs AST-based summaries similar to our approach, it lacks the bottom-up strategy and reinforcement mechanisms incorporated in \textsc{RoSum-Mcts}, leading to inferior performance. Our results confirm \textsc{RoSum-Mcts} significantly improves summarization quality across models and datasets, validating our method.

\subsection{(RQ2) Ablation Studies}
In this section, we analyze the contribution of different components of \textsc{RoSum-Mcts}, including the candidate expansion mechanism, composite reward functions, and the impact of varying reward coefficients for Verilog-Eval dataset.

\subsubsection{Impact of Expansion Mechanism (Local Vs Global)}
To evaluate the impact of the expansion mechanism, we compare three settings: (1) Full \textsc{RoSum-Mcts} using both local and global context, (2) Only local context, and (3) Only global context. Table~\ref{tab:expansion_mechanism} presents the results for GPT-4o in terms of ROUGE-L and LLM PR.

\begin{table}[h]
    \centering
    \caption{Impact of Expansion Mechanism on GPT-4o Performance}
    \label{tab:expansion_mechanism}
    \begin{tabular}{lcc}
        \toprule
        Expansion Mechanism & ROUGE-L & LLM PR \\
        \midrule
        Full \textsc{RoSum-Mcts} & \textbf{55.4} & \textbf{64.2} \\
        Only Local Context & 50.1 & 58.8 \\
        Only Global Context & 52.5 & 60.3 \\
        \bottomrule
    \end{tabular}
\end{table}

We observe that using only local context results in a significant performance drop compared to the full model, although it still outperforms the \textsc{CoDes} baseline. On the other hand, using only global context performs slightly better than the local-only variant but remains inferior to the full \textsc{RoSum-Mcts}. These findings suggest that both local and global context contribute uniquely to generating high-quality summaries, and their combination leads to the best results.

\subsubsection{Effect of Reward Functions}
To analyze the contribution of individual reward components, we conduct an ablation study by removing each reward term separately and evaluating its impact on GPT-4o and Llama-3. Table~\ref{tab:reward_ablation} presents the results for ROUGE-L and LLM PR scores.

\begin{table}[h]
    \centering
    \caption{Effect of Reward Functions on Performance}
    \label{tab:reward_ablation}
    \begin{tabular}{lcccc}
        \toprule
        Model & Setting & ROUGE-L & LLM PR \\
        \midrule
        \multirow{4}{*}{GPT-4o} & Full \textsc{RoSum-Mcts} & \textbf{55.4} & \textbf{64.2} \\
        & w/o FC & 52.3 & 60.4 \\
        & w/o LCA & 50.8 & 58.5 \\
        & w/o Fluency & 54.1 & 63.8 \\
        \midrule
        \multirow{4}{*}{Llama-3} & Full \textsc{RoSum-Mcts} & \textbf{48.8} & \textbf{55.6} \\
        & w/o FC & 44.6 & 49.7 \\
        & w/o LCA & 46.2 & 52.0 \\
        & w/o Fluency & 47.4 & 54.3 \\
        \bottomrule
    \end{tabular}
\end{table}

From the results, we find that removing the fluency reward leads to only a marginal drop in performance for both models. This suggests that modern LLMs inherently possess strong fluency and readability capabilities, making explicit fluency rewards less critical. In contrast, removing Functional Correctness (FC) or Local Content Adequacy (LCA) significantly degrades quality. GPT-4o is particularly sensitive to LCA, highlighting its importance for preserving relevant local details in summaries. Llama-3 suffers a larger performance drop without FC, showing that functional correctness is critical for its summaries. These results highlight the importance of FC and LCA in boosting quality, with fluency a secondary factor given LLMs' inherent coherence.


\subsubsection{Influence of Reward Coefficients}

\begin{figure}[h]
    \centering
    \includegraphics[width=0.5\textwidth]{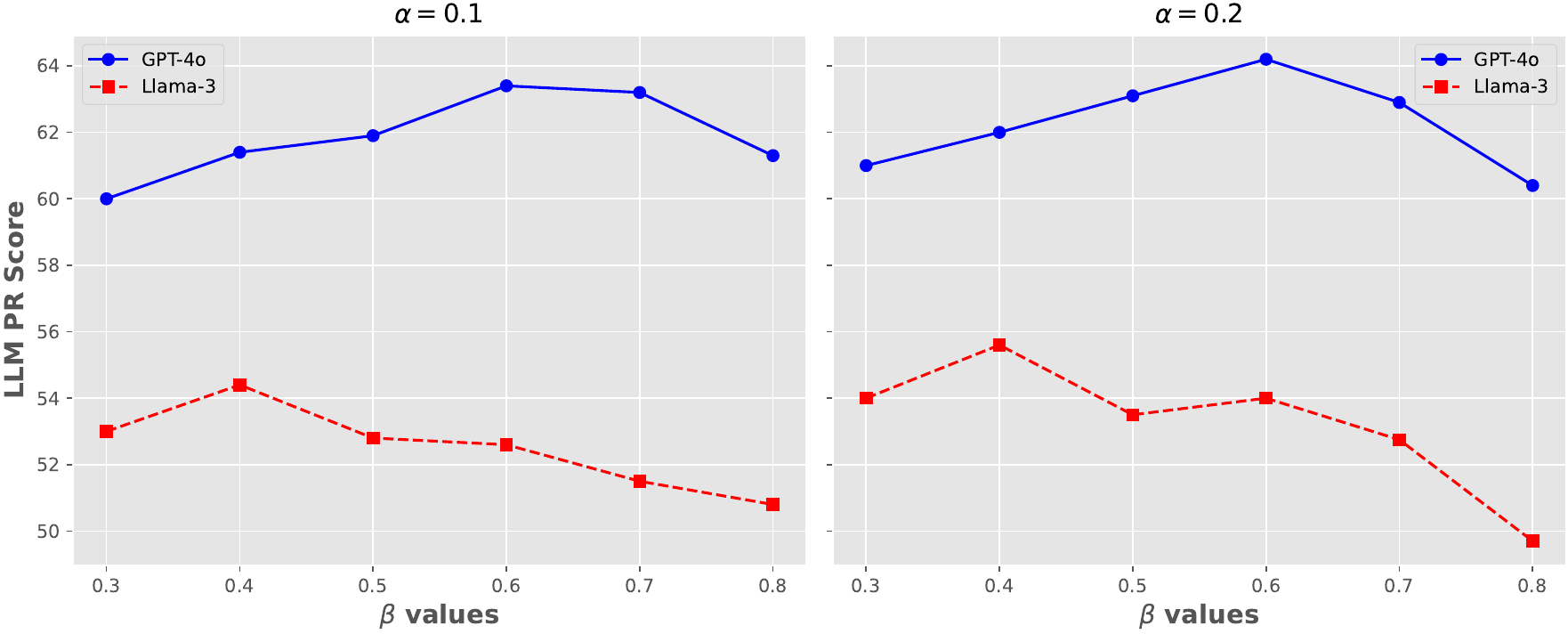}
    \caption{Effect of reward coefficients on LLM PR scores for GPT-4o and Llama-3.}
    \label{fig:coefficients_plot}
\end{figure}

Based on the previous findings, we plot line graphs for reward coefficients as $\alpha=0.1, 0.2$, varying $\beta=\{0.3,0.4,0.5,0.6,0.7,0.8\}$. This ensures that $\gamma$ is adjusted accordingly so that all coefficients sum up to 1. From the plot in Figure~\ref{fig:coefficients_plot}, we observe that GPT-4o's performance improves as $\beta$ increases, highlighting the importance of Local Content Adequacy (LCA). However, this improvement stops beyond $\beta=0.6$, indicating diminishing returns from further emphasizing LCA. On the other hand, Llama-3 exhibits greater volatility with increasing $\beta$, peaking around $\beta=0.4$, after which performance declines. This suggests that while LCA contributes positively, excessive weighting might disrupt the balance needed for effective summarization. These trends indicate that while both LCA and Functional Correctness (FC) are crucial, their relative importance varies across models. GPT-4o appears to benefit more from LCA, whereas Llama-3 shows a more delicate balance between these rewards. This reinforces the need for careful tuning of reward coefficients to optimize performance across different models.

\subsubsection{Robustness to Superficial Modifications}

\begin{figure}[h]
    \centering
    \includegraphics[width=0.6\linewidth]{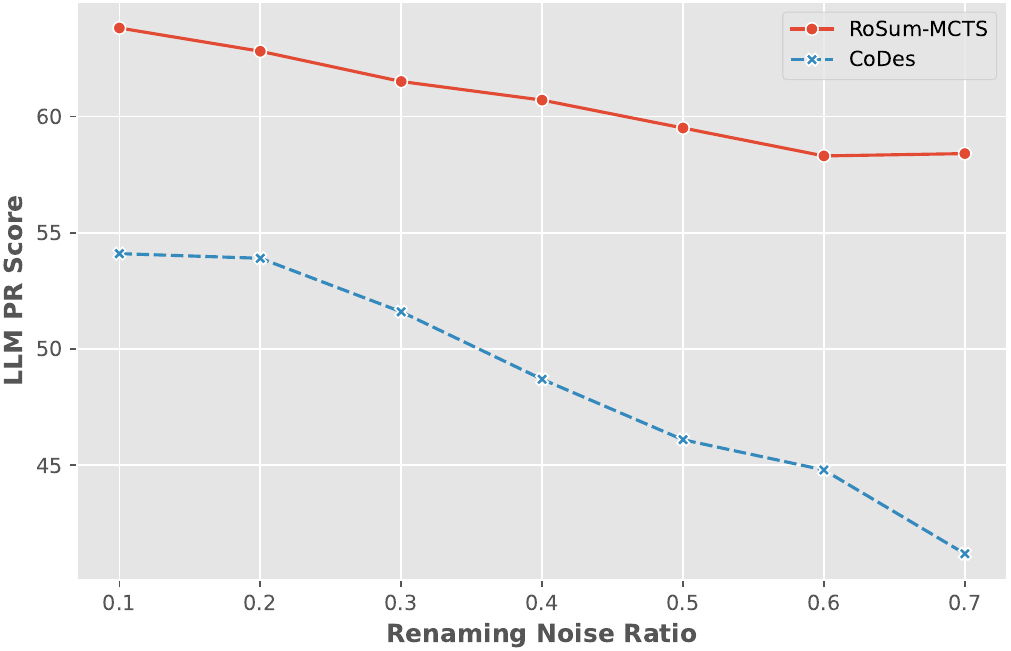}
    \caption{Effect of Renaming Noise Ratio on Summarization Robustness}
    \label{fig:renaming_robustness}
\end{figure}

To assess the robustness of our approach against superficial modifications, we introduce controlled renaming noise by systematically replacing variable names in Verilog designs with varying renaming ratios ranging from 0.1 to 0.7. We evaluate the performance of \textsc{RoSum-Mcts} and \textsc{CoDes} under these conditions using the LLM PR metric. Figure~\ref{fig:renaming_robustness} illustrates the impact of renaming noise on summarization quality. We observe that as the renaming ratio increases, \textsc{CoDes} exhibits a sharper decline in performance, indicating its sensitivity to variable renaming. In contrast, \textsc{RoSum-Mcts} maintains robust performance, with only a marginal drop before stabilizing. This suggests that our method effectively captures structural and contextual information beyond mere lexical matches, ensuring resilience against such superficial modifications.

\section{Conclusion}  

We introduced \textsc{RoSum-Mcts}, a novel code summarization approach inspired by Monte Carlo Tree Search (MCTS), leveraging structured rewards and an effective expansion mechanism. Extensive experiments across multiple LLMs and datasets demonstrate its superiority over baselines, including \textsc{CoDes}, due to its bottom-up expansion and reinforcement-driven optimization. GPT-4o achieves the best performance, followed by Llama-3 and Granite Code-34b, with Verilog summarization outperforming VHDL, likely due to greater model exposure.  Ablation studies highlight the importance of combining local and global context, with significant performance drops when using only one. \textsc{RoSum-Mcts} optimizes summaries using three key rewards: functional correctness (FC), local content adequacy (LCA), and fluency, each playing a critical role. While fluency has a marginal impact, FC and LCA are essential, with GPT-4o relying more on LCA and Llama-3 on FC. Tuning reward coefficients further reinforces the need for a balance between these factors. \textsc{RoSum-Mcts} also proves robust against renaming modifications, maintaining high summarization quality while \textsc{CoDes} degrades significantly. These findings establish \textsc{RoSum-Mcts} as an effective and resilient summarization method. Future work includes refining reward mechanisms and extending to additional languages and hardware description paradigms.

\bibliographystyle{ieeetr}
\bibliography{IEEEexample}

\begin{thebibliography}{10}

\bibitem{fakih2024llm4plc}
M.~Fakih, R.~Dharmaji, Y.~Moghaddas, G.~Quiros, O.~Ogundare, and M.~A. Al~Faruque, ``Llm4plc: Harnessing large language models for verifiable programming of plcs in industrial control systems,'' in {\em Proceedings of the 46th International Conference on Software Engineering: Software Engineering in Practice}, pp.~192--203, 2024.

\bibitem{paul2024ircoder}
I.~Paul, G.~Glava{\v{s}}, and I.~Gurevych, ``Ircoder: Intermediate representations make language models robust multilingual code generators,'' {\em arXiv preprint arXiv:2403.03894}, 2024.

\bibitem{wei2023magicoder}
Y.~Wei, Z.~Wang, J.~Liu, Y.~Ding, and L.~Zhang, ``Magicoder: Empowering code generation with oss-instruct,'' {\em arXiv preprint arXiv:2312.02120}, 2023.

\bibitem{yang2023harnessing}
Y.~Yang, S.~Xiong, A.~Payani, E.~Shareghi, and F.~Fekri, ``Harnessing the power of large language models for natural language to first-order logic translation,'' {\em arXiv preprint arXiv:2305.15541}, 2023.

\bibitem{cassano2024knowledge}
F.~Cassano, J.~Gouwar, F.~Lucchetti, C.~Schlesinger, A.~Freeman, C.~J. Anderson, M.~Q. Feldman, M.~Greenberg, A.~Jangda, and A.~Guha, ``Knowledge transfer from high-resource to low-resource programming languages for code llms,'' {\em Proceedings of the ACM on Programming Languages}, vol.~8, no.~OOPSLA2, pp.~677--708, 2024.

\bibitem{li2023starcoder}
R.~Li, L.~B. Allal, Y.~Zi, N.~Muennighoff, D.~Kocetkov, C.~Mou, M.~Marone, C.~Akiki, J.~Li, J.~Chim, {\em et~al.}, ``Starcoder: may the source be with you!,'' {\em arXiv preprint arXiv:2305.06161}, 2023.

\bibitem{roziere2023code}
B.~Roziere, J.~Gehring, F.~Gloeckle, S.~Sootla, I.~Gat, X.~E. Tan, Y.~Adi, J.~Liu, R.~Sauvestre, T.~Remez, {\em et~al.}, ``Code llama: Open foundation models for code,'' {\em arXiv preprint arXiv:2308.12950}, 2023.

\bibitem{zhao2024codev}
Y.~Zhao, D.~Huang, C.~Li, P.~Jin, Z.~Nan, T.~Ma, L.~Qi, Y.~Pan, Z.~Zhang, R.~Zhang, {\em et~al.}, ``Codev: Empowering llms for verilog generation through multi-level summarization,'' {\em arXiv preprint arXiv:2407.10424}, 2024.

\bibitem{cui2024origen}
F.~Cui, C.~Yin, K.~Zhou, Y.~Xiao, G.~Sun, Q.~Xu, Q.~Guo, D.~Song, D.~Lin, X.~Zhang, {\em et~al.}, ``Origen: Enhancing rtl code generation with code-to-code augmentation and self-reflection,'' {\em arXiv preprint arXiv:2407.16237}, 2024.

\bibitem{gao2024autovcoder}
M.~Gao, J.~Zhao, Z.~Lin, W.~Ding, X.~Hou, Y.~Feng, C.~Li, and M.~Guo, ``Autovcoder: A systematic framework for automated verilog code generation using llms,'' in {\em 2024 IEEE 42nd International Conference on Computer Design (ICCD)}, pp.~162--169, IEEE, 2024.

\bibitem{pei2024betterv}
Z.~Pei, H.-L. Zhen, M.~Yuan, Y.~Huang, and B.~Yu, ``Betterv: Controlled verilog generation with discriminative guidance,'' {\em arXiv preprint arXiv:2402.03375}, 2024.

\bibitem{thakur2023autochip}
S.~Thakur, J.~Blocklove, H.~Pearce, B.~Tan, S.~Garg, and R.~Karri, ``Autochip: Automating hdl generation using llm feedback,'' {\em arXiv preprint arXiv:2311.04887}, 2023.

\bibitem{sun2024classification}
W.~Sun, B.~Li, G.~L. Zhang, X.~Yin, C.~Zhuo, and U.~Schlichtmann, ``Classification-based automatic hdl code generation using llms,'' {\em arXiv preprint arXiv:2407.18326}, 2024.

\bibitem{vijayaraghavan2024chain}
P.~Vijayaraghavan, A.~Nitsure, C.~Mackin, L.~Shi, S.~Ambrogio, A.~Haran, V.~Paruthi, A.~Elzein, D.~Coops, D.~Beymer, {\em et~al.}, ``Chain-of-descriptions: Improving code llms for vhdl code generation and summarization,'' in {\em Proceedings of the 2024 ACM/IEEE International Symposium on Machine Learning for CAD}, pp.~1--10, 2024.

\bibitem{li2024rethinkmcts}
Q.~Li, W.~Xia, K.~Du, X.~Dai, R.~Tang, Y.~Wang, Y.~Yu, and W.~Zhang, ``Rethinkmcts: Refining erroneous thoughts in monte carlo tree search for code generation,'' {\em arXiv preprint arXiv:2409.09584}, 2024.

\bibitem{dainese2024generating}
N.~Dainese, M.~Merler, M.~Alakuijala, and P.~Marttinen, ``Generating code world models with large language models guided by monte carlo tree search,'' {\em arXiv preprint arXiv:2405.15383}, 2024.

\bibitem{browne2012survey}
C.~B. Browne, E.~Powley, D.~Whitehouse, S.~M. Lucas, P.~I. Cowling, P.~Rohlfshagen, S.~Tavener, D.~Perez, S.~Samothrakis, and S.~Colton, ``A survey of monte carlo tree search methods,'' {\em IEEE Transactions on Computational Intelligence and AI in games}, vol.~4, no.~1, pp.~1--43, 2012.

\bibitem{liu2023verilogeval}
M.~Liu, N.~Pinckney, B.~Khailany, and H.~Ren, ``Verilogeval: Evaluating large language models for verilog code generation,'' in {\em 2023 IEEE/ACM International Conference on Computer Aided Design (ICCAD)}, pp.~1--8, IEEE, 2023.

\bibitem{vijayaraghavan2024vhdl}
P.~Vijayaraghavan, L.~Shi, S.~Ambrogio, C.~Mackin, A.~Nitsure, D.~Beymer, and E.~Degan, ``Vhdl-eval: A framework for evaluating large language models in vhdl code generation,'' in {\em 2024 IEEE LLM Aided Design Workshop (LAD)}, pp.~1--6, IEEE, 2024.

\bibitem{sun2024source}
W.~Sun, Y.~Miao, Y.~Li, H.~Zhang, C.~Fang, Y.~Liu, G.~Deng, Y.~Liu, and Z.~Chen, ``Source code summarization in the era of large language models,'' {\em arXiv preprint arXiv:2407.07959}, 2024.

\bibitem{Takamaeda:2015:ARC:Pyverilog}
S.~Takamaeda-Yamazaki, ``Pyverilog: A python-based hardware design processing toolkit for verilog hdl,'' in {\em Applied Reconfigurable Computing}, vol.~9040 of {\em Lecture Notes in Computer Science}, pp.~451--460, Springer International Publishing, Apr 2015.

\bibitem{pyVHDLParser}
P.~Lehmann, ``pyvhdlparser: A vhdl parser written in python,'' 2019.

\bibitem{gao2021simcse}
T.~Gao, X.~Yao, and D.~Chen, ``Simcse: Simple contrastive learning of sentence embeddings,'' in {\em International Conference on Learning Representations}, 2021.

\bibitem{Lin:2004:ROUGE}
C.-Y. Lin, ``Rouge: A package for automatic evaluation of summaries,'' in {\em Text Summarization Branches Out}, 2004.

\bibitem{yuan2023evaluating}
Z.~Yuan, J.~Liu, Q.~Zi, M.~Liu, X.~Peng, and Y.~Lou, ``Evaluating instruction-tuned large language models on code comprehension and generation,'' {\em arXiv preprint arXiv:2308.01240}, 2023.

\bibitem{paszke2019pytorch}
A.~Paszke, S.~Gross, F.~Massa, A.~Lerer, J.~Bradbury, G.~Chanan, T.~Killeen, Z.~Lin, N.~Gimelshein, L.~Antiga, {\em et~al.}, ``Pytorch: An imperative style, high-performance deep learning library,'' {\em Advances in Neural Information Processing Systems}, vol.~32, 2019.

\bibitem{openai2023chatgpt}
OpenAI, ``Chatgpt api.'' \url{https://openai.com/blog/chatgpt/}, 2023.

\end{thebibliography}
\newpage
\appendix

\section{Prompt Templates for Candidate Generation}

\begin{table}[htbp]
\centering
\caption{Prompt templates used for candidate generation in \textsc{RoSum-Mcts}. Here, $N_c=|node.children|$, denotes the number of child nodes for the current AST node.}
\label{tab:prompt_variants}
\begin{tabular}{|p{2.5cm}|p{5cm}|}
\hline
\textbf{Variation Type} & \textbf{Prompt Template} \\
\hline
Local Summary Context & \texttt{Combine the following child summaries to form a concise summary:} [$S_{child_1}, S_{child_1}, .. S_{N_c}$]\\
\hline
Local Inference & \texttt{Infer the higher-level functionality based solely on the following child summaries:} [$S_{child_1}, S_{child_1}, .. S_{N_c}$] \\
\hline
Local with Global Code Context & \texttt{Infer the functionality of the current node based on the following intermediate steps (child summaries) in the broader context of the full code:} \newline \texttt{Child Summaries:} [$S_{child_1}, S_{child_1}, .. S_{N_c}$] \newline \texttt{Full Code: \{global\_code\_text\}} \\
\hline
Local with Global Summary Context & \texttt{Infer the functionality of the current node using the following intermediate steps (child summaries) in the broader context of the overall code summary:} \newline \texttt{Child Summaries:} [$S_{child_1}, S_{child_1}, .. S_{N_c}$] \newline \texttt{Global Code Summary: \{global\_code\_summary\}} \\
\hline
\end{tabular}
\end{table}

\section{Implementation Details and Computational Cost}
\subsection{Implementation Details}
The \textsc{RoSum-Mcts} framework is implemented in Python, leveraging the capabilities of modern deep learning frameworks such as PyTorch or TensorFlow for neural model computations. The system integrates:
\begin{itemize}
    \item \textbf{HDL Parsing:} Utilizing \texttt{pyverilog} and \texttt{pyVHDLParser} for AST construction.
    \item \textbf{LLM-based Summarization:} Pre-trained language models (e.g., GPT variants) are used to generate both the preliminary global summary and the candidate summaries at each node.
    \item \textbf{Candidate Generation and Evaluation:} Multiple prompt templates are employed to generate candidate summaries, while reward evaluation leverages state-of-the-art semantic embedding models such as SimCSE and prompt-based LLM scoring.
\end{itemize}
The design allows for parallel processing at several stages, notably during candidate generation and reward computation, to mitigate latency introduced by LLM calls.

\subsection{Computational Cost and Scalability}
Although integrating AST parsing, LLM-based summarization, candidate generation, and reward evaluation may seem computationally heavy, \textsc{RoSum-Mcts} remains efficient and scalable in practice. Our experiments report an average AST size of $\sim{62.7}$ nodes per HDL module. Moreover, the average prompt size across our evaluation datasets (VerilogEval and VHDL-Eval) is $\sim{290}$ tokens, making LLM-based inference cost-effective for both cloud APIs and local deployments. While HDL code is often sparsely commented, our datasets comprises of examples without any inline comments to assess generalizability. We observe that \textsc{RoSum-Mcts}, due to its AST-level and semantic reward structure, is able to generate better summaries compared to baseline approaches. We deploy on two NVIDIA V100 GPUs (32GB each), utilizing high-throughput inference and embedding computation. To ensure scalability and robustness, especially when extending to real-world designs, we integrate the following strategies:

\noindent \textbf{Batch Processing.} Candidate generation and embedding computations are batched across nodes, significantly reducing per-call overhead and ensuring GPU utilization is optimized—especially beneficial for local models with low throughput or proprietary models with cost constraints.

\noindent \textbf{Efficient Embedding Computation.} We use parallelized inference of SimCSE-like models for semantic similarity scoring during reward computation, enabling fast candidate evaluation.

\noindent \textbf{Controlled Tree Traversal \& AST Pruning.} With $\sim{62.7}$ nodes per module, we avoid exhaustive exploration by pruning low-reward subtrees based on reward heuristics, which reduces the number of LLM calls without compromising summarization quality.

\noindent \textbf{AST Chunking for Large Modules.} While current benchmarks may involve mid-sized HDL modules, we simulate scalability to larger designs via chunked traversal. For larger designs, the AST can be partitioned and processed independently, with summary integration handled via global context anchoring. This enables practical deployment for real-world industrial modules.

\noindent \textbf{Asynchronous Execution.} LLM calls and reward evaluations are asynchronously scheduled, maximizing hardware resource utilization and minimizing wait times.

\noindent \textbf{Handling LLM Variability.} To account for the known stochasticity of LLM outputs, we report aggregate metrics over multiple sampling runs for a subset of modules and observe consistent ranking among methods. While full multiple-run evaluation is computationally expensive, especially when operating over tree-structured search spaces, we validate that our hierarchical selection framework dampens variance across runs due to the reward-guided selection of summaries.

\noindent \textbf{Local vs Proprietary Model Defense.} Proprietary APIs incur per-call or per-token costs, whereas local models incur fixed compute costs but higher latency per interaction. Our batching, pruning, and chunking strategies reduce total calls and token volume—minimizing monetary cost for API users and improving throughput for self-hosted setups.

These combined strategies ensure that \textsc{RoSum-Mcts} handles mid-scale ASTs ($\sim{62.7 }$ nodes, $\sim{290}$ token prompts) with both computational efficiency and cost-effectiveness—across diverse deployment environments.

\end{document}